\documentclass{article}

\usepackage[final]{nips_2016}

\usepackage[utf8]{inputenc} 
\usepackage[T1]{fontenc}    
\usepackage{hyperref}       
\usepackage{url}            
\usepackage{booktabs}       
\usepackage{amsfonts}       
\usepackage{nicefrac}       
\usepackage{microtype}      
\usepackage[table]{xcolor}
\usepackage[toc,page]{appendix}

\usepackage{cancel,multirow,graphicx,caption,subcaption,pifont,tikz} 
\usetikzlibrary{shapes,arrows,matrix,positioning,backgrounds,through,shadows,bayesnet,calc,decorations.pathmorphing}

\def \GRID {1.75cm}

\definecolor{dynamicscolor}{rgb}{.6,0,0}
\definecolor{correctioncolor}{rgb}{0,0,.6}
\definecolor{emissioncolor}{rgb}{0,.5,0}

\tikzstyle{latent} = [circle,fill=white,draw=black,inner sep=1pt,minimum size=25pt, font=\fontsize{10}{10}\selectfont]

\tikzstyle{os} = [densely dashed,circle,fill=white,draw=black,inner sep=1pt,minimum size=25pt, font=\fontsize{10}{10}\selectfont]

\tikzstyle{observed} = [circle,fill=gray!25,draw=black,inner sep=1pt,minimum size=25pt, font=\fontsize{10}{10}\selectfont]

\tikzstyle{target} = [circle,fill=gray!25,draw=black,inner sep=1pt,minimum size=25pt, font=\fontsize{10}{10}\selectfont]

\tikzstyle{input} = [circle,fill=gray!25,draw=black,inner sep=1pt,minimum size=25pt, font=\fontsize{10}{10}\selectfont]

\tikzstyle{criterion} = [decoration={snake,segment length=1.5mm,amplitude=.5mm},decorate,fill=white,line width=1pt,shorten >=0pt]

\tikzstyle{dynamics}=[-stealth, line width=1.5pt,color=dynamicscolor]
\tikzstyle{shoot}=[ densely dashed]
\tikzstyle{correction}=[-stealth, line width=1.5pt, color=correctioncolor]
\tikzstyle{emission}=[-latex', line width=1.5pt, color=emissioncolor]
\tikzstyle{densely dashed}= [dash pattern=on 3pt off .7pt]
\tikzstyle{separator}=[densely dashed,line width=1pt,draw=black!30]

\title{Connecting Generative Adversarial Networks\\ and Actor-Critic Methods}

\author{
  David Pfau, Oriol Vinyals \\
  Google DeepMind\\
  \texttt{\{pfau,vinyals\}@google.com} \\
}

\begin{document}

\maketitle

\begin{abstract}
Both generative adversarial networks (GAN) in unsupervised learning and actor-critic methods in reinforcement learning (RL) have gained a reputation for being difficult to optimize. Practitioners in both fields have amassed a large number of strategies to mitigate these instabilities and improve training. Here we show that GANs can be viewed as actor-critic methods in an environment where the actor cannot affect the reward. We review the strategies for stabilizing training for each class of models, both those that generalize between the two and those that are particular to that model. We also review a number of extensions to GANs and RL algorithms with even more complicated information flow. We hope that by highlighting this formal connection we will encourage both GAN and RL communities to develop general, scalable, and stable algorithms for multilevel optimization with deep networks, and to draw inspiration across communities.
\end{abstract}

\section{Introduction}
\label{sec:intro}

Most problems in machine learning are formulated as an optimization problem over a single objective. However, a number of problems in machine learning lack a single unified cost, and instead consist of a hybrid of several models, each of which passes information to other models but tries to minimize its own private loss function. This upsets many of the assumptions behind most learning algorithms, and applying ordinary methods like gradient descent often leads to pathological behavior such as oscillations or collapse onto degenerate solutions. Despite these practical difficulties, there is great potential in models with hybrid or multilevel losses, and it has been hypothesized that the combination of many different local losses underlies the functioning of the brain as well \cite{marblestone2016towards}.

Actor-critic methods (AC) \cite{sutton1999policy,konda2003onactor} and generative adversarial networks (GANs) \cite{goodfellow2014generative} are two such classes of multilevel optimization problems which have close parallels. In both cases the information flow is a simple feedforward pass from one model which either takes an action (AC) or generates a sample (GANs) to a second model which evaluates the output of the first model. In both cases, the second model is the only one which has direct access to special information in the environment, either reward information (AC) or real samples from the distribution in question (GANs), and the first model must learn based on error signals from the second model alone. GANs and AC methods have important differences as well, and Section~\ref{sec:algo} we review both classes of algorithms and show a construction that bridges the two. Both of these models suffer from stability issues, and techniques for stabilizing training have been developed largely independently by the two communities. In Section~\ref{sec:stable} we review the different methods used to address model pathologies. We also provide a review of other related methods in the supplemental materials. The aim of this note is to raise awareness of the strong connection between the two classes of models.

\section{Algorithms}
\label{sec:algo}
Both GANs and AC can be seen as bilevel or two-time-scale optimization problems, where one model is optimized with respect to the optimum of another model:

\begin{eqnarray}
x^* & = & \arg\min_{x\in \mathcal{X}} F(x,y^*(x)) \\
y^*(x) & = & \arg\min_{y \in \mathcal{Y}} f(x,y)
\end{eqnarray}

Bilevel optimization problems have been extensively studied in operations research, including the study of actor-critic methods \cite{konda2002actor}, but largely in the setting where the two optimization problems are linear or convex programs \cite{colson2007overview}. By contrast, the recent applications we review here have largely used deep neural networks as the class of functions over which to optimize.

\subsection{Generative Adversarial Networks}
\label{sec:gan}

Generative adversarial networks \cite{goodfellow2014generative} formulate the unsupervised learning problem as a game between two opponents - a generator $G$ which samples from a distribution, and a discriminator $D$ which classifies the samples as real or false. Typically the generator is represented as a deterministic feedforward neural network through which a fixed noise source $z \sim \mathcal{N}(0,I)$ is passed and the discriminator is another neural network which maps an image to a binary classification probability. The GAN game is then formulated as a zero sum game where the value is the cross-entropy loss between the discriminator's prediction and the true identity of the image as real or generated, which we denote $y$:

\begin{eqnarray}
    \min_G \max_D \mathbb{E}_{w,y}[y \mathrm{log}D(w) + (1-y)\mathrm{log}(1-D(w))] & = & \nonumber \\
    \min_G \max_D \mathbb{E}_{w\sim p_\mathrm{data}}[\mathrm{log}D(w)] + \mathbb{E}_{z \sim \mathcal{N}(0,I)}[\mathrm{log}(1-D(G(z)))] & &
    \label{eqn:gan}
\end{eqnarray}

To make sure the generator has gradients from which to learn even when the discriminator's classification accuracy is high, the generator's loss function is usually formulated as maximizing the probability of classifying a sample as true rather than minimizing its probability of being classified false. The modified loss is still easily formulated as a bilevel optimization problem:
\begin{eqnarray}
F(D,G) & = & -\mathbb{E}_{w\sim p_\mathrm{data}}[\mathrm{log}D(w)] - \mathbb{E}_{z \sim \mathcal{N}(0,I)}[\mathrm{log}(1-D(G(z)))] \\
f(D,G) & = & -\mathbb{E}_{z\sim \mathcal{N}(0,I)}[\mathrm{log}D(G(z))]
\label{eqn:gan2}
\end{eqnarray}

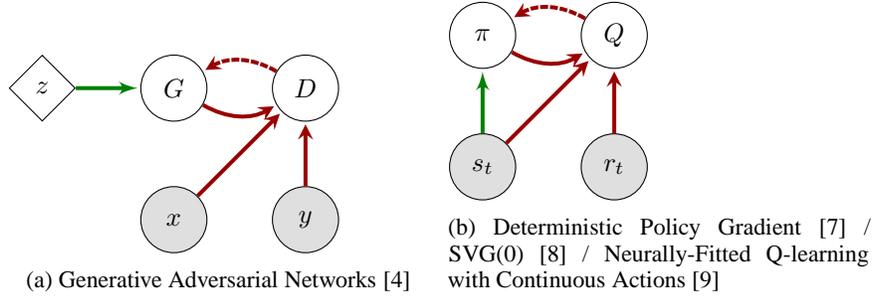
\begin{figure}
    \centering
    \begin{subfigure}[b]{0.4\textwidth}
        {
        \begin{tikzpicture}[node distance=\GRID, shorten >=1pt]

\node[det] (z) {$z$};
\node[latent] (G)[right of=z]{$G$};
\node[latent] (D)[right of=G]{$D$};
\node[obs] (x)[below of=G]{$x$};
\node[obs] (y)[below of=D]{$y$};

\draw[emission](z) -- (G);
\draw[dynamics](G) to[bend right] (D);
\draw[dynamics](x) -- (D);
\draw[dynamics](y) -- (D);
\draw[dynamics,shoot] (D) to[bend right] (G){};	

\end{tikzpicture}
        }
        \caption{Generative Adversarial Networks \cite{goodfellow2014generative}}
        \label{fig:gan}
    \end{subfigure}
    ~
    \begin{subfigure}[b]{0.4\textwidth}
        {
        \begin{tikzpicture}[node distance=\GRID, shorten >=1pt]

\node[latent] (pi){$\pi$};
\node[obs] (s)[below of=pi]{$s_t$};
\node[latent] (Q)[right of=pi]{$Q$};
\node[obs] (r)[below of=Q]{$r_t$};

\draw[emission](s) -- (pi);
\draw[dynamics](s) -- (Q);
\draw[dynamics](pi) to[bend right] (Q);
\draw[dynamics](r) -- (Q);
\draw[dynamics,shoot] (Q) to[bend right] (pi){};

\end{tikzpicture}
        }
        \caption{Deterministic Policy Gradient \cite{silver2014determinisitc} / SVG(0) \cite{heess2015learning} / Neurally-Fitted Q-learning with Continuous Actions \cite{hafner2011reinforcement}}
        \label{fig:dpg}
    \end{subfigure}
    \caption{Information structure of GANs and AC methods. Empty circles represent models with a distinct loss function. Filled circles represent information from the environment. Diamonds represent fixed functions, both deterministic and stochastic. Solid lines represent the flow of information, while dotted lines represent the flow of gradients used by another model. Paths which are analogous between the two models are highlighted in red. The dependence of $Q$ on future states and the dependence of future states on $\pi$ are omitted for clarity.}
    \label{fig:comparison}
\end{figure}

\subsection{Actor-Critic Methods}
\label{sec:ac}

Actor-critic methods are a long-established class of techniques in reinforcement learning \cite{sutton1999policy}. While most reinforcement learning algorithms either focus on learning a value function, like value iteration and TD-learning, or learning a policy directly, as in policy gradient methods, AC methods learn both simultaneously - the actor being the policy and the critic being the value function. In some AC methods, the critic provides a lower-variance baseline for policy gradient methods than estimating the value from returns. In this case even a bad estimate of the value function can be useful, as the policy gradient will be unbiased no matter what baseline is used. In other AC methods, the policy is updated with respect to the approximate value function, in which case pathologies similar to those in GANs can result. If the policy is optimized with respect to an incorrect value function, it may lead to a bad policy which never fully explores the space, preventing a good value function from being found and leading to degenerate solutions. A number of techniques exist to remedy this problem.

Formally, consider the typical MDP setting for RL, where we have a set of states $\mathcal{S}$, actions $\mathcal{A}$, a distribution over initial states $p_0(s)$, transition function $\mathcal{P}(s_{t+1}|s_t, a_t)$, reward distribution $\mathcal{R}(s_t)$ and discount factor $\gamma\in [0,1]$. The aim of actor-critic methods is to simultaneously learn an action-value function $Q^{\pi}(s, a)$ that predicts the expected discounted reward:

\begin{equation}
Q^{\pi}(s, a) = \mathbb{E}_{s_{t+k}\sim\mathcal{P},r_{t+k}\sim\mathcal{R},a_{t+k}\sim\pi}\left[\sum_{k=1}^{\infty} \gamma^k r_{t+k} \middle| s_t = s, a_t = a\right]
\label{eqn:bellman}
\end{equation}
and learn a policy that is optimal for that value function: 

\begin{equation}
\pi^* = \arg\max_{\pi} \mathbb{E}_{s_0\sim p_0, a_0\sim\pi}[Q^\pi(s_0,a_0)]
\label{eqn:policy}
\end{equation}
We can express $Q^{\pi}$ as the solution to a minimization problem:
\begin{equation}
Q^\pi = \arg\min_Q \mathbb{E}_{s_t, a_t \sim\pi}[\mathcal{D}(\mathbb{E}_{s_{t+1},r_t,a_{t+1}}[r_t + \gamma Q(s_{t+1}, a_{t+1})]||Q(s_t, a_t))]
\label{eqn:bellman-residual}
\end{equation}
Where $\mathcal{D}(\cdot||\cdot)$ is any divergence that is positive except when the two are equal. Now the actor-critic problem can be expressed as a bilevel optimization problem as well:

\begin{eqnarray}
F(Q,\pi) & = & \mathbb{E}_{s_t, a_t \sim\pi}[\mathcal{D}(\mathbb{E}_{s_{t+1},r_t,a_{t+1}}[r_t + \gamma Q(s_{t+1}, a_{t+1})]||Q(s_t, a_t))] \\
f(Q,\pi) & = & -\mathbb{E}_{s_0\sim p_0, a_0\sim\pi}[Q^\pi(s_0,a_0)]
\label{eqn:ac-bilevel}
\end{eqnarray}

There are many AC methods that attempt to solve this problem. Traditional AC methods optimize the policy through policy gradients and scale the policy gradient by the TD error, while the action-value function is updated by ordinary TD learning. We focus on deterministic policy gradients (DPG) \cite{silver2014determinisitc, lillicrap2015continuous} and its extension to stochastic policies, SVG(0) \cite{heess2015learning}, as well as neurally-fitted Q-learning with continuous actions (NFQCA) \cite{hafner2011reinforcement}. These algorithms are all intended for the case where actions and observations are continuous, and use neural networks for function approximation for both the action-value function and policy. This is an established approach in RL with continuous actions \cite{prokhorov1997adaptive}, and all methods update the policy by passing back gradients of the estimated value with respect to the actions rather than passing the TD error directly. The distinction between the methods lies mainly in the way training proceeds. In NFQCA, the actor and critic are trained in batch mode after every episode, while in DPG and SVG(0), networks are trained online using temporal difference updates.

\subsection{GANs as a kind of Actor-Critic}
\label{sec:connection}

The similarities between GANs and AC methods are summarized in Figure~\ref{fig:comparison}. In both situations, one model has access to information about errors from the environment (the discriminator in GANs and the critic in AC), while the other model must be updated based only on gradient information from the first model.

We can make this connection more precise and describe an MDP in which GANs are a modified type of actor-critic method. Consider an MDP where the actions set every pixel in an image. The environment randomly chooses either to show the image the actor generates or show a real image. Let the reward from the environment be 1 if the environment chose the real image and 0 if not. This MDP is stateless as the image generated by the actor does not affect future data.

An actor-critic architecture learning in this environment clearly closely resembles the GAN game. A few adjustments have to be made to make it identical. If the actor had access to the state it could trivially pass a real image forward, so the actor must be a \textit{blind} actor, with no knowledge of the state. Since the MDP is stateless this doesn't preclude the actor from learning. While the mean-squared Bellman residual is usually used as a loss for the critic, cross-entropy should be used instead to match the GAN loss. This still gives a loss function with a consistent value function as the minimum. Since the actor receives gradients of the value rather than gradients of the Bellman residual in RL, a scaling term proportional to $\frac{{\partial} \mathcal{D}}{\partial Q}$, where $\mathcal{D}$ is the cross-entropy, should be included in the gradients passed to the actor (though in practice, a different loss is used for the generator precisely to deal with complications from this scaling term). Lastly, the actor's parameters should not be updated if the environment shows a real image. This could be done by having the critic zero its gradients with respect to the action if the reward is 1. Thus GANs can be seen as a modified actor-critic method with blind actors in stateless MDPs.

There are several peculiar aspects of this MDP that give rise to some behaviors not typically associated with actor-critic methods. First, it is not immediately obvious why an actor-critic algorithm should lead to adversarial behavior. Typically the actor and critic are trying to optimize complimentary loss functions, rather than optimize the same loss in different directions. The adversarial behavior in GANs comes about because the MDP in which the GAN game is played is one in which the actor cannot have any causal effect on the reward. Essentially, it is an MDP in which the true policy gradient is always zero. A critic, however, cannot learn the causal structure of the game from input examples alone, and moves in the direction of features that predict reward. The actor moves in a direction to increase reward based on the best estimate from the critic, but this change cannot lead to an increase in the true reward, so the critic will quickly learn to assign lower value in the direction the actor has moved. Thus the updates to the actor and critic, which ideally would be orthogonal (as in compatible actor-critic) instead becomes adversarial. It is also worth noting that there are important consequences to the environment being only partially observable to the actor. For fully observable MDPs, the optimal policy is always deterministic. For GANs, however, the generator matching the true distribution is a fixed point of the minimax problem. Despite these differences between GANs and typical RL problems, we believe there are enough similarities to merit investigation into techniques to improve training that generalize between the two settings.





\section{Stabilizing Strategies}
\label{sec:stable}

\begin{table}
\begin{center}
 \begin{tabular}{||c c c||} 
 \hline
 Method & GANs & AC \\ [0.5ex] 
 \hline\hline
 Freezing learning & \cellcolor{green!50}yes & \cellcolor{green!50}yes \\
 \hline
 Label smoothing & \cellcolor{green!50}yes & \cellcolor{yellow!50}no \\
 \hline
 Historical averaging &  \cellcolor{green!50}yes & \cellcolor{yellow!50}no   \\ 
 \hline
 Minibatch discrimination & \cellcolor{green!50}yes & \cellcolor{yellow!50}no \\
 \hline
 Batch normalization & \cellcolor{green!50}yes & \cellcolor{green!50}yes   \\
 \hline
 Target networks & \cellcolor{red!50}n/a  & \cellcolor{green!50}yes \\ 
 \hline
 Replay buffers & \cellcolor{yellow!50}no & \cellcolor{green!50}yes \\
 \hline
 Entropy regularization & \cellcolor{yellow!50}no & \cellcolor{green!50}yes \\ 
 \hline
 Compatibility & \cellcolor{yellow!50}no & \cellcolor{green!50}yes \\[1ex] 
 \hline
\end{tabular}
\caption{Summary of different approaches used to stabilize and improve training for GANs and AC methods. Those approaches that have been shown to improve performance are in green, those that have not yet been demonstrated to improve training are in yellow, and those that are not applicable to the particular method are in red.}
\label{table:tricks}
\end{center}
\end{table}

Having reviewed the basics of GANs, actor-critic algorithms and their extensions, here we discuss the ``tricks of the trade" used by each community to make them work. The different methods are summarized in Table~\ref{table:tricks}. While not meant as an exhaustive list, we have included those that we believe have either made the largest impact in their fields or have the greatest potential for crossover between fields.

\subsection{GANs}

\begin{enumerate}
    \item \textbf{Freezing learning}. Frequently either the generator or discriminator will outpace the other and the GAN will get stuck in a degenerate solution. One simple remedy is to freeze learning in one model when it begins to get too strong. A very similar approach has been taken successfully in actor-critic learning, where learning is frozen for either the actor or the critic if the magnitude of the TD error goes below or above a certain threshold \cite{degrispersonal}. 
    \item \textbf{Label smoothing}. A simple trick to prevent gradients from vanishing when the discriminator's predictions are very confident, label smoothing replaces 0/1 labels with $\epsilon$/$1-\epsilon$, which guarantees the generator will always have informative gradients. While specific to classification, there's no reason this couldn't equally well be applied in an RL setting where the rewards are 0/1 and gradients of the critic vanish.
    \item \textbf{Historical averaging} \cite{salimans2016improved}. Loosely inspired by fictitious play in game theory, historical averaging adds a drag term to gradient descent that penalizes steps that are too far from the past average over parameters. Even though the average over parameters doesn't have an intuitive meaning, this method is effective at preventing oscillations due to two models optimizing different objectives. Historical averaging is also closely related to Polyak-Ruppert averaging and extensions \cite{polyak1990new,ruppert1988efficient}. While Polyak-Ruppert style averaging has been formally analyzed by those in the RL community \cite{konda2004convergence}, it has not been adopted as a standard ``tool of the trade" to the best of our knowledge. The use of replay buffers in DPG \cite{lillicrap2015continuous} is also conceptually similar to fictitious play, but only applicable to the actor.
    \item \textbf{Minibatch discrimination} \cite{salimans2016improved}. To prevent the generator from collapsing onto a single sample, minibatch discrimination extends the role of the discriminator from classifying single images to classifying entire minibatches. This helps increase the entropy of samples from the generator. In RL, the analogous problem of underexploration has been addressed by adding a penalty to the actor that encourages higher entropy policies. Computing the entropy directly becomes impractical for complex stochastic policies, and an RL equivalent to minibatch discrimination could be an effective alternative for encouraging exploration in continuous spaces.
    \item \textbf{Batch normalization} \cite{radford2016unsupervised, salimans2016improved}. Batch normalization has been critical for scaling GANs to deep convolutional networks, and virtual batch normalization has extended this by using a constant reference batch to prevent correlations in predictions due to the other elements of the minibatch. Batch normalization has also been helpful in a large fraction of environments DPG has been tried for, and it has not been investigated whether virtual batch normalization leads to even greater improvement.
\end{enumerate}

\subsection{Actor-Critic}

\begin{enumerate}
    \item \textbf{Replay buffers} \cite{lillicrap2015continuous}. Replay buffers have been very effective at removing correlations from the training data in both discrete and continuous RL settings. They are also conceptually similar to fictitious play in game theory, where one player plays the best response to the average over policies of the other player. Unlike fictitious play, where both players play against the historical average opponent, in the AC setting replay buffers can only be used for the critic. The critic can make off-policy updates based on actions chosen by the actor in the past, but the actor cannot learn from gradients from the critic in the past, since those gradients were taken with respect to different actions. Similarly for GANs, we can keep a buffer of previously generated images to prevent the discriminator from fitting too closely to the current generator. We have experimented with using a replay buffer for GANs, but were not able to generate asymptotically correct samples even for simple distributions.
    \item \textbf{Target networks} \cite{lillicrap2015continuous}. Since the action value function appears twice in the Bellman recursion, stability can be an issue in Q-learning with function approximation. Target networks address this by fixing one of the networks in TD updates. Since the GAN game can be seen as a stateless MDP, the second appearance of the action value function disappears and learning the discriminator becomes an ordinary regression problem. For this reason we don't consider target networks applicable to the GAN setting. However other multilevel optimization problems with Q-learning as a subproblem would likely benefit. It's also worth noting that self-contrastive estimation \cite{goodfellow2014distinguishability} is conceptually similar to target networks, so the idea of a target network can still be applicable to other density estimation problems.
    \item \textbf{Entropy regularization}. Actor-critic methods often fail to sufficiently explore the action space. To address this, an additional reward is sometimes added that encourages high-entropy policies. When the action space is discrete this is easily accomplished, but is somewhat more difficult in the continuous control setting. Notably, an analogous problem is often encountered with GANs, where the generator collapses onto sampling only a few modes. Any method for encouraging exploration in continuous control is likely transferable to increasing sample diversity in GANs.
    \item \textbf{Compatibility}. One of the unique theoretical developments of actor-critic methods is the notion of a \textit{compatible} critic. If the critic is a linear function of the gradient of the policy with respect to its parameters, then using this in place of the true value in the gradient of the expected return gives an unbiased approximation if the critic is optimal \cite{sutton1999policy}. This compatible critic is also closely related to the natural gradient of the expected return, and can be used for efficient natural gradient descent on the policy \cite{kakade2001natural}. While elegant, it is not clear if the notion of compatibility can be naturally extended to the GAN setting. Since the true value of any policy will always be 0.5 in the GAN MDP, the true policy gradient is always zero. We would generally prefer our GANs to be adversarial than compatible.
\end{enumerate}

\section{Conclusions}

Combining deep learning with multilevel optimization holds great promise for a diverse array of problems in machine learning and AI. Already GANs and actor-critic methods have made large impacts on their respective fields, despite the inherent difficulties in optimization and exploration. We hope that by pointing out the deep connections between the two we encourage the development and adoption of general techniques and free flow of ideas between different communities.

\subsubsection*{Acknowledgments}

Thanks to David Silver, Thomas Degris, Marc Lanctot, Nicholas Heess, Ian Goodfellow, Scott Reed, Josh Merel, Razvan Pascanu, Greg Wayne, Shakir Mohamed, Balaji Lakshminarayanan, Chelsea Finn and Sergey Levine for helpful discussions and to Yuval Tassa for help with the figures.

\small

\bibliographystyle{unsrt}   
\bibliography{biblio}

\normalsize

\section*{Supplemental Material}

\renewcommand{\thesubsection}{\Alph{subsection}}
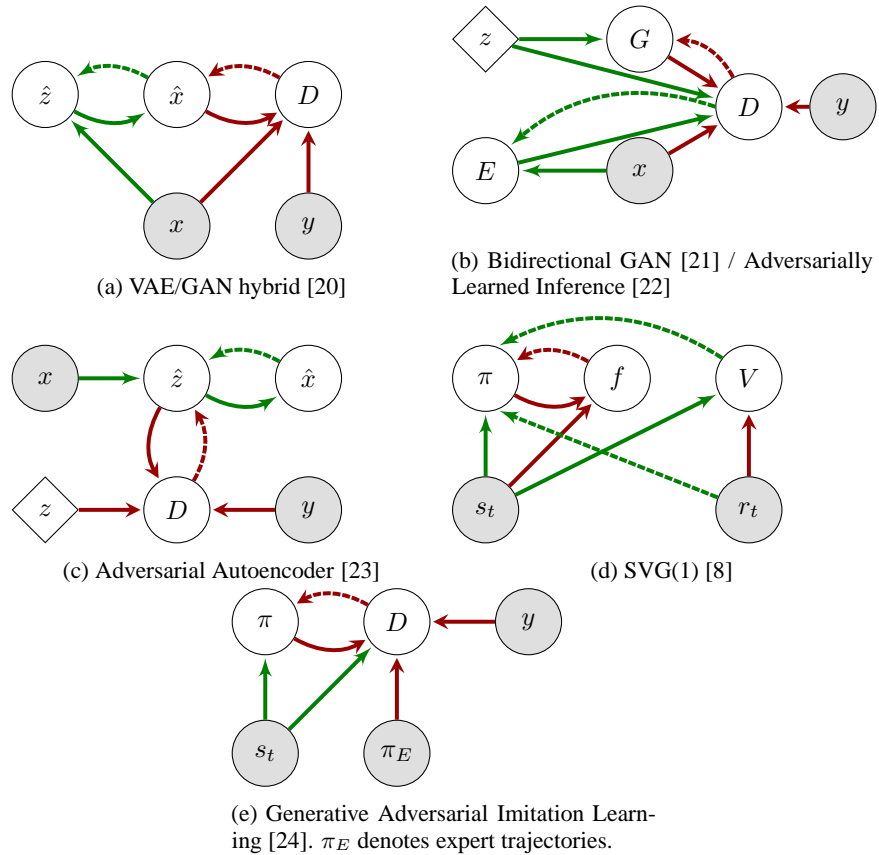
\begin{figure}[h]
    \centering
    \begin{subfigure}[b]{0.4\textwidth}
        {
        \begin{tikzpicture}[node distance=\GRID, shorten >=1pt]

\node[latent] (q){$\hat{z}$};
\node[latent] (G)[right of=q]{$\hat{x}$};
\node[latent] (D)[right of=G]{$D$};
\node[obs] (x)[below of=G]{$x$};
\node[obs] (y)[below of=D]{$y$};

\draw[emission](x) -- (q);
\draw[emission](q) to[bend right] (G);
\draw[dynamics](G) to[bend right] (D);
\draw[dynamics](x) -- (D);
\draw[dynamics](y) -- (D);
\draw[dynamics,shoot] (D) to[bend right] (G){};	
\draw[emission,shoot] (G) to[bend right] (q);

\end{tikzpicture}
        }
        \caption{VAE/GAN hybrid \cite{larsen2015autoencoding}}
        \label{fig:vaegan}
    \end{subfigure}
    ~
    \begin{subfigure}[b]{0.4\textwidth}
        {
        \begin{tikzpicture}[node distance=\GRID, shorten >=1pt]

\node[det] (z) {$z$};
\node[latent] (G)[right of=z, xshift=3mm]{$G$};
\node[latent] (D)[right of=G, yshift=-9mm, xshift=-3mm]{$D$};
\node[obs] (x)[below of=G]{$x$};
\node[obs] (y)[right of=D, xshift=-5mm]{$y$};
\node[latent] (E) [below of=z]{$E$};

\draw[emission](z) -- (G);
\draw[dynamics](G) -- (D);
\draw[dynamics](x) -- (D);
\draw[dynamics](y) -- (D);
\draw[emission](x) -- (E);
\draw[emission](E) -- (D);
\draw[emission,shoot](node cs:name=D,angle=180) to[bend right] (E);
\draw[emission](z) -- (D);
\draw[dynamics,shoot] (D) to[bend right] (G){};	

\end{tikzpicture}
        }
        \caption{Bidirectional GAN \cite{donahue2016adversarial} / Adversarially Learned Inference \cite{dumoulin2016adversarially}}
        \label{fig:bigan}
    \end{subfigure}
    ~
    \begin{subfigure}[b]{0.4\textwidth}
        {
        \begin{tikzpicture}[node distance=\GRID, shorten >=1pt]

\node[obs] (x) {$x$};
\node[latent] (q) [right of=x]{$\hat{z}$};
\node[det] (p) [below of=x]{$z$};
\node[latent] (pq) [right of=q]{$\hat{x}$};
\node[latent] (D) [right of=p]{$D$};
\node[obs] (y) [right of=D]{$y$};

\draw[emission] (x) -- (q);
\draw[emission] (q) to[bend right] (pq);
\draw[dynamics] (q) to[bend right] (D);
\draw[dynamics] (p) -- (D);
\draw[dynamics] (y) -- (D);

\draw[dynamics,shoot] (D) to [bend right] (q);
\draw[emission,shoot] (pq) to [bend right] (q);



\end{tikzpicture}
        }
        \caption{Adversarial Autoencoder \cite{makhzani2015adversarial}}
        \label{fig:aae}
    \end{subfigure}
    ~
    \begin{subfigure}[b]{0.4\textwidth}
        {
        \begin{tikzpicture}[node distance=\GRID, shorten >=1pt]

\node[latent] (pi){$\pi$};
\node[latent] (f)[right of=pi]{$f$};
\node[obs] (s)[below of=pi]{$s_t$};
\node[latent] (V)[right of=f]{$V$};
\node[obs] (r)[below of=V]{$r_t$};

\draw[emission](s) -- (pi);
\draw[dynamics](s) -- (f);
\draw[emission](s) -- (V);
\draw[dynamics](pi) to[bend right] (f);
\draw[dynamics](r) -- (V);
\draw[emission,shoot] (V) to[bend right] (node cs:name=pi,angle=70);
\draw[dynamics,shoot] (f) to[bend right] (pi);
\draw[emission,shoot] (r) -- (node cs:name=pi,angle=-70);

\end{tikzpicture}
        }
        \caption{SVG(1) \cite{heess2015learning}}
        \label{fig:svg}
    \end{subfigure}
    ~
    \begin{subfigure}[b]{0.4\textwidth}
        {
        \begin{tikzpicture}[node distance=\GRID, shorten >=1pt]


\node[latent] (G){$\pi$};
\node[latent] (D)[right of=G]{$D$};
\node[obs] (s)[below of=G]{$s_t$};
\node[obs] (x)[below of=D]{$\pi_E$};
\node[obs] (y)[right of=D]{$y$};

\draw[emission](s) -- (G);
\draw[emission](s) -- (D);
\draw[dynamics](G) to[bend right] (D);
\draw[dynamics](x) -- (D);
\draw[dynamics](y) -- (D);
\draw[dynamics,shoot] (D) to[bend right] (G){};	

\end{tikzpicture}
        }
        \caption{Generative Adversarial Imitation Learning \cite{ho2016generative}. $\pi_E$ denotes expert trajectories.}
        \label{fig:gail}
    \end{subfigure}
    \caption{Other machine learning problems which can be framed as multilevel optimization problems.}
\end{figure}
Beyond the basic GAN and AC methods described in Section~\ref{sec:algo}, there have been many extensions and other problems in machine learning which have even greater complexity. We review some of them here.

\subsection{GAN extensions}
There have been many extensions to GANs since the original publication. In \cite{nowozin2016f} it was shown that GANs can be interpreted as minimizing a lower bound on the f-divergence between the generator distribution and true distribution:

\begin{equation}
    \min_G \max_{T\in\mathcal{T}} \mathbb{E}_{w\sim p_\mathrm{data}}[T(w)] - \mathbb{E}_{z \sim \mathcal{N}(0,I)}[f^*(T(G(z)))] \le \min_G D_f(P_\mathrm{data}||Q_G)
    \label{eqn:f-gan}
\end{equation}
Where $Q_G$ is the distribution induced by the generator network $G$. For the appropriate choice of convex $f$ the original GAN loss is recovered as a bound on the Jensen-Shannon divergence. Other choices of $f$ lead to lower bounds on other divergences such as the KL divergence or squared Hellinger distance. GANs were also reformulated in \cite{zhao2016energy} by replacing the discriminator with an energy-based autoencoder trained with a contrastive loss. Both f-GANs and energy-based GANs preserve the information and gradient flow from the original GAN formulation but change the loss function.

GANs lack a natural way to infer the latent state for new data, leading to a number of extensions that include an inference network in addition to the generator and discriminator. A hybrid variational autoencoder \cite{kingma2013auto, rezende2014stochastic} and GAN architecture was trained in \cite{larsen2015autoencoding}, while \cite{dumoulin2016adversarially} and \cite{donahue2016adversarial} both proposed an extension to GANs which trained the discriminator to classify the latent state of true data induced by an inference network in addition to the generated images. The structure of these models is illustrated in Figures~\ref{fig:vaegan} and \ref{fig:bigan} respectively. These extensions all added a third model to the mix, complicating optimization even further.

Further extensions include adversarial autoencoders \cite{makhzani2015adversarial} which adversarially learn the structure of the approximate posterior in variational autoencoders (shown in Figure~\ref{fig:aae}), and InfoGANs \cite{chen2016infogan}, which augment the latent state in the generative model with another set of units trained to maximize the mutual information with the data to learn the most important factors of variation in the model.

\subsection{AC extensions}

DPG, SVG(0) and NFQCA are not the only AC methods that work with continuous actions. Asynchronous advantage actor-critic (A3C) has shown promise for RL with both discrete and continuous actions \cite{mnih2016asynchronous}. A3C learns only a state value function $V(s_t)$ rather than an action value function $Q(s_t, a_t)$ and thus cannot pass back gradients of the value with respect to the action to the actor. Instead it approximates the action value with the rewards from several steps of experience and passes the TD error to the actor. It is therefore less closely connected to GANs, however it is worth pointing out that it has been a successful alternative to DPG and NFQCA as an actor-critic algorithm for continuous control.

Stochastic value gradients (SVG) \cite{heess2015learning} generalize DPG to stochastic policies in a number of ways, giving a spectrum from model-based to model-free algorithms. While SVG(0) is a direct stochastic generalization of DPG, SVG(1) combines an actor, critic and model $f$. The actor is trained through a combination of gradients from the critic, model and reward simultaneously.

\subsection{Imitation Learning and Inverse Reinforcement Learning}
Another machine learning problem closely related to GANs and actor-critic methods is that of learning control policies from expert trajectories, variously referred to as apprenticeship learning \cite{abbeel2005exploration}, imitation learning or inverse reinforcement learning \cite{ng2000algorithms} depending on the exact approach taken.

It has long been known that imitation learning can be formulated as a minimax problem where the aim is to learn the cost function that minimizes the amount by which the optimal policy outperforms the expert policy \cite{syed2007game}. In \cite{ho2016generative}, a deep connection was shown between imitation learning and GANs. Learning a cost function was shown to be equivalent to minimizing the regularized distance between state-action occupancy distributions for the expert policy and learned policy. By using a deep neural network with the cross-entropy loss as the distance between occupancy distributions, learning a cost directly could be circumvented and the inverse reinforcement learning problem was cast in nearly the same form as GAN learning. The only major difference from GANs is that instead of learning to generate images by gradient descent a policy is learned by trust region policy optimization \cite{schulman2015trust}. In principle an actor-critic method could be substituted for direct policy optimization, and the imitation learning problem would become a three-level optimization problem that contains both GANs and actor-critic as subproblems.

It has also been shown in \cite{finn2016connection} that generative adversarial networks have a close connection with maximum entropy inverse reinforcement learning. In particular, when the density function of the generative model is known, the GAN objective is identical to the MaxEnt IRL objective, and GAN training is identical to guided cost learning, which approximates the partition function in MaxEnt IRL using a learned importance sampling estimator. While in this paper we have shown a connection between actor-critic learning in a {\em particular} MDP and all GANs, their work shows that a particular extension to GANs is the same as guided cost learning in {\em all} cases.



\end{document}